\documentclass[letterpaper]{article}
\usepackage[left=2cm,right=2cm,top=2cm,bottom=2cm]{geometry}
\usepackage[affil-it]{authblk}

\providecommand{\keywords}[1]{\textbf{\textit{Keywords: }} #1}

\usepackage{times}

\usepackage{epstopdf}
\usepackage{mathtools}
\usepackage{amsmath,amsfonts,amssymb,amsthm}
\usepackage{graphicx}
\usepackage{subcaption}
\usepackage[font=scriptsize,labelfont=bf,labelsep=space]{caption}
\usepackage{placeins}
\usepackage{float}
\usepackage{pgfplots}
\usepackage{enumerate}
\usepackage[utf8]{inputenc}
\usepackage[T1]{fontenc}
\usepackage{lmodern}
\usepackage{hyperref}
\usepackage{isomath}
\usepackage{xcolor}
\usepackage{algorithm}
\usepackage[algo2e]{algorithm2e}
\usepackage{algpseudocode}
\usepackage{xpatch}
\mathtoolsset{showonlyrefs}

\usepackage{array}
\newcolumntype{P}[1]{>{\centering\arraybackslash}p{#1}}

\algnewcommand\algorithmicinput{\textbf{Input:}}
\algnewcommand\Input{\item[\algorithmicinput]}
\algnewcommand\algorithmicoutput{\textbf{Output:}}
\algnewcommand\Output{\item[\algorithmicoutput]}
\algnewcommand\algorithmicgiven{\textbf{Given}}
\algnewcommand\Given{\item[\algorithmicgiven]}
\algnewcommand\algorithmicchoose{\textbf{Choose}}
\algnewcommand\Choose{\item[\algorithmicchoose]}

\algdef{SE}[SUBALG]{Indent}{EndIndent}{}{\algorithmicend\ }%
\algtext*{Indent}
\algtext*{EndIndent}

\makeatletter
\xpatchcmd{\algorithmic}{\itemsep\z@}{\itemsep=0.8ex}{}{}
\makeatother

\newcommand{\N}{\mathbb N}
\newcommand{\PP}{\mathcal P}
\newcommand{\CC}{\mathcal C}
\newcommand{\QQ}{\mathcal Q}
\renewcommand{\vec}[1]{\mathbf{#1}}

\newtheorem{theorem}{Theorem}[section]

\newtheorem{cor}{Corollary}[section]
\newtheorem{rem}{Remark}[section]

\renewcommand{\qed}{$\blacksquare$}
\def\hindu{\arabic}

\renewcommand{\theequation}{\hindu{section}.\hindu{equation}}
\def\RR{{\mathbb R}}

\def\ZZ{{\mathbb Z}}

\def\x{\mathbf{x}}

\def\y{\mathbf{y}}

\def\z{\mathbf{z}}

\def\be{\begin{equation}}
\def\ee{\end{equation}}
\def\bea{\begin{eqnarray}}
\def\eea{\end{eqnarray}}

\def\disp{\displaystyle}

\def\donchitre#1#2{\vskip 6.5cm\noindent
\parbox[t]{1in}{\special{eps:#1.eps x=6.5cm y=5.5cm}}
\hbox to 7cm{}\parbox[t]{0.0cm}{\special{eps:#2.eps x=6.5cm y=5.5cm}}}

\def\XX{{\mathbb X}}
\def\BB{{\mathbb B}}

\usepackage{tikz}
\usetikzlibrary{shapes.geometric, shapes.multipart, shapes, arrows, shadings,fit,calc}
\tikzstyle{blank} = []
\tikzstyle{block} = [rectangle, rounded corners, minimum width=6em, minimum height=5em,text centered, draw=black]
\tikzstyle{line} = [draw, -latex']
\tikzstyle{arrow} = [->,>=stealth]
\tikzstyle{cloud} = [draw, ellipse, node distance=3cm,minimum height=2em]

\graphicspath{{figures/}, {tables/}}

\title{A function approximation approach to the prediction of blood glucose levels}
	\author[1]{H.N. Mhaskar
	\thanks{Email: hrushikesh.mhaskar@cgu.edu}}
	\affil[1]{Institute of Mathematical Sciences, Claremont Graduate University, Claremont, CA, 91711 USA}
	\author[2]{S.V. Pereverzyev}
	\affil[2]{The Johann Radon Institute for Computational and Applied Mathematics (RICAM), Austrian Academy of Sciences, Linz, Austria.}
	\author[3]{M.D. van der Walt
		\thanks{Email: mvanderwalt@westmont.edu}}
	\affil[3]{Department of Mathematics and Computer Science, Westmont College, Santa Barbara, CA, USA}
	
	\date{}

\begin{document}

\maketitle

\begin{abstract}
The problem of real time prediction of blood glucose (BG) levels based on the readings from a continuous glucose monitoring (CGM) device is a problem of great importance in diabetes care, and therefore, has attracted a lot of research in recent years, especially based on machine learning. 
An accurate prediction with a 30, 60, or 90 minute prediction horizon has the potential of saving millions of dollars in emergency care costs.
In this paper, we treat the problem as one of function approximation, where the value of the BG level at time $t+h$ (where $h$ the prediction horizon) is considered to be an unknown function of $d$ readings prior to the time $t$. 
This unknown function may be supported in particular on some unknown submanifold of the $d$-dimensional Euclidean space.
While manifold learning is classically done in a semi-supervised setting, where the entire data has to be known in advance, we use recent ideas to achieve an accurate function approximation in a supervised setting; i.e., construct a model for the target function.
We use the state-of-the-art clinically relevant PRED-EGA grid to evaluate our results, and demonstrate that for a real life dataset, our method performs better than a standard deep network, especially in hypoglycemic and hyperglycemic regimes.
One noteworthy aspect of this work is that the training data and test data may come from different distributions.

\noindent \keywords{{supervised learning, Hermite polynomial, continuous glucose monitoring, blood glucose prediction, prediction error-grid analysis.}} 
\end{abstract}

\section{Introduction}
\label{section-intro}

Diabetes is a major disease affecting humanity. According to the 2020 National Diabetes Statistics report \cite{centers2020national}, more than 34,000,000 people had diabetes in the United States alone in 2018,  contributing to  nearly 270,000 deaths and costing nearly 327 billion dollars in 2017.
There is so far no cure for diabetes, and one of the important ingredients in keeping it in control is to monitor blood glucose levels regularly.
Continuous glucose monitoring (CGM) devices are used increasingly for this purpose.
These devices typically record blood sugar readings at 5 minute intervals, resulting in a time series. 
This paper aims at predicting the level and rate of change of the sugar levels at a medium range prediction horizon; i.e., looking ahead for 30-60 minutes.
A clinically reliable prediction of this nature is extremely useful in conjunction with other communication devices in avoiding unnecessary costs and dangers. 
For example, hypoglycemia may lead to loss of consciousness, confusion or seizures \cite{mujahid2021machine}. 
However, since the onset of hypoglycemic periods are often silent (without indicating symptoms), it can be very hard for a human to predict in advance. 
If these predictions are getting communicated to a hospital by some wearable communication device linked with the CGM device, then a nurse could alert the patient if the sugar levels are expected to reach dangerous levels, so that the patient could drive to the hospital if necessary, avoiding a dangerous situation, not to mention an expensive call for an ambulance.\\

Naturally, there are many papers dealing with prediction of blood glucose (BG) based not just on CGM recordings but also other factors such as meals, exercise, etc. 
In particular, machine learning has been used extensively for this purpose. 
We refer to \cite{mujahid2021machine} for a recent review of BG prediction methods based on machine learning, with specific focus on predicting and detecting hypoglycemia.\\

The fundamental problem of machine learning is the following. 
We have (training) data  of the form $\{(\x_j,f(\x_j)+\epsilon_j)\}$, where, for some integer $d\ge 1$, $\x_j\in\RR^d$ are realizations of a random variable,  $f$ is an unknown real valued function, and the $\epsilon_j$'s are realizations of a mean zero random variable. 
The distributions of both the random variables are not known.
The problem is to approximate $f$ with a suitable model, especially for the points which are not in the training data.
For example, in the prediction of a time series $x_0,x_1,\ldots$ with a prediction horizon of $h$ based on $d$ past readings can be (and is in this paper) viewed as a problem of approximating an unknown function $f$ such that $x_{j+h}=f(x_{j-d+1}, \ldots, x_j)$ for each $j$.
Thus, $\x_j=(x_{j-d+1}, \ldots, x_j)$ and $y_j=x_{j+h}$. \\

Such problems have been studied in approximation theory for more than a century.
Unfortunately, classical approximation theory techniques do not apply directly in this context, mainly because the data does not necessarily become dense on any known domain such as a cube, sphere, etc., and we cannot control the locations of the points $\x_j$.
This is already clear from the example given above -- it is impossible to require that the BG readings be at a pre-chosen set of points.\\

Manifold learning tries to ameliorate this problem by assuming that the data lies on some unknown manifold, and developing techniques to learn various quantities related to the manifold, in particular, the eigen-decomposition of the Laplace-Beltrami operator. 
Since these objects must be learnt from the data, these techniques are applicable in the semi-supervised setting, in which we have all the data points $\x_j$ available in advance.
In \cite{mhas_sergei_maryke_diabetes2017}, we have used deep manifold learning to predict BG levels using CGM data. 
Being in the semi-supervised setting, this approach cannot be used directly for prediction based on data that  was not  included in the original data set we worked with. \\

Recently, we have discovered \cite{mhaskar2019deep} a more direct method to approximate functions on unknown manifolds without trying to learn anything about the manifold itself, and giving an approximation with theoretically guaranteed errors on the entire manifold; not just the points in the original data set.
The purpose of this paper is to use this new tool for BG prediction based on CGM readings which can be used on patients not in the same data sets, and for readings for the patients in the data set which are not included among the training data. \\

In this context, the numerical accuracy of the prediction alone is not clinically relevant.
The Clarke Error-Grid Analysis (C-EGA) \cite{clarke1987evaluating} is devised to predict whether the prediction is reasonably accurate, or erroneous without serious consequences (clinically uncritical), or results in unnecessary treatment (overcorrection), or dangerous (either because it fails to identify a rise or drop in BG or because it confuses a rise in BG for a drop or vice versa). The Prediction Error-Grid Analysis (PRED-EGA), on the other hand, categorizes a prediction as accurate, erroneous with benign (not serious) consequences or erroneous with potentially dangerous consequences in the hypoglycemic (low BG), euglycemic (normal BG), and hyperglycemic (high BG) ranges separately. This stratification is of great importance because consequences caused by a prediction error in the hypoglycemic range are very different from ones in the euglycemic or the hyperglycemic range. In addition, the categorization is based not only on reference BG estimates paired with the BG estimates predicted for the same moments (as C-EGA does), but also on reference BG directions and rates of change paired with the corresponding estimates predicted for the same moments. It is argued in \cite{sivananthan2011assessment} that this is a much better clinically meaningful assessment tool than C-EGA, and is therefore the assessment tool we choose to use in this paper. \\
 
Since different papers on the subject use different criterion for assessment it is impossible for us to compare our results with those in most of these papers.
Besides, a major objective of this paper is to illustrate the use and utility of the theory in \cite{mhaskar2019deep} from the point of view of machine learning.
So, we will demonstrate the superiority of our results in the hypoglycemic and hyperglycemic regimes over those obtained by a blind training of a deep network using the MATLAB Deep Learning Toolbox.  We will also compare our results with those obtained by the techniques used in \cite{naumova2012meta, mhaskar2013filtered, mhas_sergei_maryke_diabetes2017}. 
However, we note that  these methods are not directly comparable to ours. In \cite{naumova2012meta}, the authors use the data for only one patient as training data, which leads to a meta-learning model that can be used serendipitously on other patients, although a little bit of further training is still required for each patient to apply this model.  
In contrast, we require a more extensive ``training data'', but there is no training in the classical sense, and the  application to new patients consists of a simple matrix vector multiplication.
The method in \cite{mhaskar2013filtered} is a linear extrapolation method, where the coefficients are learnt separately on each patient. 
It does not result in a model that can be trained once for all and applied to other patients in a simple manner.
As pointed out earlier, the method in \cite{mhas_sergei_maryke_diabetes2017} requires that the entire data set be available even before the method can start to work. 
So, unlike our method, it cannot be trained on a part of the data set and applied to a completely different data set, not available at the beginning.
\\

We will explain the problem and the evaluation criterion in Section~\ref{section-problem}. 
Prior work on this problem is reviewed briefly in Section~\ref{section-priorwork}. 
The methodology and algorithm used in this paper are described in Section~\ref{section-method}. 
The results are discussed in Section~\ref{section-results}. 
The mathematical background behind the methods described in Section~\ref{section-method} is summarized in Appendix~\ref{hermitesect}.


\section{Prior work}
\label{section-priorwork}

Given the importance of the problem, many researchers have worked on it in several directions.  Below we highlight the main trends relevant to our work.\\

The majority of machine learning models (30\% of those studied in the survey paper \cite{mujahid2021machine}) are based on artificial neural networks such as convolutional neural networks, recurrent neural networks and deep learning techniques. For example, in \cite{zhu2020dilated}, the authors develop a deep learning model based on a dilated recurrent neural network to achieve 30-minute blood glucose prediction. The authors assess the accuracy of their method by calculating the root mean squared error (RMSE) and mean absolute relative difference (MARD). While they achieve good results, a limitation of the method is the reliance on various input parameters such as meal intake and insulin dose which are often recorded manually and might therefore be inaccurate. In \cite{pappada2011neural}, on the other hand, a feed-forward neural network is designed with eleven neurons in the input layer (corresponding to variables such as CGM data, the rate of change of glucose levels, meal intake and insulin dosage), and nine neurons with hyperbolic tangent transfer function in the hidden layer. The network was trained with the use of data from 17 patients and tested on data from 10 other patients for a 75-minute prediction, and evaluated using C-EGA. Although good results are achieved in the C-EGA grid in this paper, a limitation of the method is again the large amount of additional input information necessary to design the model, as described above.\\

A second BG prediction class employs decision trees. The authors of \cite{seo2019machine}, for example, use random forests to perform a 30-minute prediction to specifically detect postprandial hypoglycemia, that is, hypoglycemia occurring after a meal. They use statistical analyses like sensitivity and specificity to evaluate their method. In \cite{vahedi2018predicting}, prediction is again achieved by random forests and incorporating a wide variety of additional input features such as physiological and physical activity parameters. Mean Absolute Percentage Error (MAPE) is used as assessment methodology. The authors of \cite{jung2017prediction} use a classification and regression tree (CART) to perform a 15-minute prediction using BG data only, again using sensitivity and specificity to test the performance of their method. \\

A third class of methods employ kernel-based regularization techniques to achieve prediction (for example, \cite{naumova2012meta} and references therein), where Tikhonov regularization is used to find the best least square fit to the data, assuming the minimizer belongs to a reproducing kernel Hilbert space (RKHS). Of course, these methods are quite sensitive to the choice of kernel and regularization parameters. Therefore, the authors in \cite{naumova2012meta} develop methods to choose both the kernel and regularization parameter adaptively, or through meta-learning (``learning to learn'') approaches. C-EGA and PRED-EGA are used as performance metrics. \\

Time series techniques are also employed in the literature. In \cite{reifman2007predictive}, a tenth-order auto-regression (AR) model is developed, where the AR coefficients are determined through a regularized least square method. The model is trained patient-by-patient, typically using the first 30\% of the patient's BG measurements, for a 30-minute or 60-minute prediction. The method is tested on a time series containing glucose values measured every minute, and evaluation is again done through the C-EGA grid.
The authors in \cite{sparacino2007glucose} develop a first-order AR model, patient-by-patient, with time-varying AR coefficients determined through weighted least squares. Their method is tested on a time series containing glucose values measured every three minutes, and quantified using statistical metrics such as RMSE. As noted in \cite{naumova2012meta}, these methods seem to be sensitive to gaps in the input data.\\

A summary of the methods described above is given in Table \ref{table-lit}.\\

It should be noted that the majority of BG prediction models (63.64\% of those analyzed in \cite{mujahid2021machine}), including many of the references described above, are dependent on additional input features (such as meal intake, insulin dosage and physical activity) in addition to historical BG data. Since such data could be inaccurate and hard to obtain, we have intentionally made the design decision in our work to base our method on historical BG data only.

\begin{table}
	\begin{center}
		\begin{tabular}{cccP{60mm}c}
			\hline
			Ref. & Method & PH (min) & Data & Assessment \\ \hline
			\cite{zhu2020dilated} & DL & 30 & BG, meal intake, insulin dose & RMSE, MARD \\\hline 
			\cite{pappada2011neural} & NN & 75 & BG, BG rate of change, meal intake, insulin dose & C-EGA \\ \hline 
			\cite{seo2019machine} & RF & 30 & BG & sensitivity / specificity \\ \hline 
			\cite{vahedi2018predicting} & RF & 60 & BG, physical activity, physiological measurements, meal intake & MAPE \\ \hline 
			\cite{jung2017prediction} & CART & 15 & BG & sensitivity / specificity \\ \hline 
			\cite{naumova2012meta} & KBR & 30, 60, 75 & BG & C-EGA \\ 
			&  & 10, 20 & BG & PRED-EGA \\ \hline 
			\cite{reifman2007predictive} & AR & 30, 60 & BG & C-EGA \\ \hline 
			\cite{sparacino2007glucose} & AR & 30 & BG & RMSE \\ \hline 
		\end{tabular}
	\caption{A summary of related work on BG prediction. PH = prediction horizon, DL = deep learning, NN = neural network, RF = random forest, CART = classification and regresssion tree, KBR = kernel-based regularization, AR = autoregression. RMSE = root mean squared error, MARD = mean absolute relative difference, C-EGA = Clarke error-grid analysis, MAPE = mean absolute percentage error, PRED-EGA = prediction error-grid analysis.}
	\label{table-lit}
	\end{center}
\end{table}

\section{The set-up}
\label{section-problem}

We use two different clinical data sets provided by the DirectNet Central Laboratory \cite{DirecNet2005}, which lists BG levels of different patients taken at 5-minute intervals with the CGM device; i.e., for each patient $p$ in the patient set $P$, we are given a time series $\{s_p(t_j)\}$, where $s_p(t_j)$ denotes the BG level at time $t_j$. 
The relevant details of the two data sets are described in Table~\ref{table-data}.

\begin{center}
	\begin{table}[H]
		\begin{center}
			\begin{tabular}{c|cc}
				\hline
				& Dataset D & Dataset J \\ \hline 
				No of patients & 25 & 25 \\ \hline 
				Total no of observations & 5542 & 5837 \\ \hline
				Avg no of observations per patient & 221.68 & 233.48 \\ \hline
				Min BG & 40 & 48.07 \\ \hline
				Max BG & 356 & 390.93 \\ \hline
				Mean BG & 128.15 & 163.92 \\ \hline
				SD in BG & 60.61 & 67.59 \\ \hline
				Avg \% hypoglycemic readings per patient & 13.50 & 4.33 \\ \hline
				Avg \% euglycemic readings per patient & 69.83 & 57.98 \\ \hline
				Avg \% hyperglycemic readings per patient & 16.67 & 37.69 \\ \hline
			\end{tabular}
			\caption{Summary statistics for datasets D and J}
			\label{table-data}
		\end{center}
	\end{table}
	
\end{center}

Our goal is to predict for each $j$, the level $s_p(t_{j+h})$, given readings $s_p(t_j),\ldots, s_p(t_{j-d+1})$ for appropriate values of $h$ and $d$.  
We took $d=7$ (a sampling horizon $t_{j} - t_{j-d+1}$ of 30 minutes has been suggested as the optimal one for BG prediction in \cite{mhaskar2013filtered,hayes2009algorithm}). We tested prediction horizons of 30 minutes ($h=6$) (the most common prediction horizon according to the analysis in \cite{mujahid2021machine}), 60 minutes ($h=12$) and 90 minutes ($h=18$).  \\

\section{Methodology in the current paper}
\label{section-method}

\subsection{Data reformulation}
For each patient $p$ in the set $P$ of patients in the data sets, the data is in the form of a time series $\left\{ s_p(t_j) \right\}_{j=1}^{N_p}$ of BG levels at time $t_j$, where $t_{j}-t_{j-1} = 5$ minutes.
We re-organize the data in the form
$$
\mathcal{P}^*=\left\{\left(\left( s_p(t_{j-d+1}), \cdots, s_p(t_j) \right), s_p(t_{j+h})\right) : j=d,\cdots,N_p-h, \ p\in P \right\}.
$$
For any patient $p\in P$, we will abbreviate
$$
\vec{x}_{p,j} = \left( s_p(t_{j-d+1}), \cdots, s_p(t_j) \right), \qquad y_{p,j}=s_p(t_{j+h}),
$$
and write 
$$ \PP := \left\{ \vec{x}_{p,j} = \left( s_p(t_{j-d+1}), \cdots, s_p(t_j) \right) :  j=d,\cdots,N_p-h, \ p\in P \right\}. $$

The problem of BG prediction is then seen as the classical problem of machine learning; i.e., to find a functional relationship $f$ such that $f(\x_{p,j})\approx y_{p,j}$, for all $p$ and $ j=d,\cdots,N_p-h$. 

\subsection{Training data selection} 
\label{subsection-training}

To obtain training data, we form the training  set $C$ by randomly selecting (according to a uniform probability distribution) $c$\% of the patients in $P$. The training data are now defined to be all the data of each patient in $C$, that is,
$$ \CC^{\star} := \left\{ (\vec{x}_{p,j}, y_{p,j}) = \left( \left( s_p(t_{j-d+1}), \cdots, s_p(t_j) \right), s_p(t_{j+h}) \right) : j=d,\cdots,N_p-h, \ p\in C \right\}.$$
Since we define the training and test data in terms of
patients, choosing the entire data for a patient in the training (respectively, test) data, we may now simplify the notation by writing $(\x_j,y_j)$ rather than $(\x_{p,j},y_{p,j})$, with the understanding that if $\x_j$ represents a data point for a patient $p$, $y_j$ is the corresponding value for the same patient $p$. 
For future reference, we also define
$$ \CC := \left\{ \vec{x}_j = \left( s_p(t_{j-d+1}), \cdots, s_p(t_j) \right) : p\in C \right\} $$ 
and the testing patient set $Q=P \setminus C$, with
$$ \QQ := \PP\setminus \CC = \left\{ \vec{x}_j = \left( s_p(t_{j-d+1}), \cdots, s_p(t_j) \right) : p\in Q \right\}. $$

\subsection{BG range classification} 
\label{subsection-classification}

To get accurate results in each BG range, we want to be able to adapt the training data and parameters used for predictions in each range -- as noted previously, consequences of prediction errors in the separate BG ranges are very different. To this end, we divide the measurements in $\CC$ into three clusters $\CC_o, \CC_e$ and $\CC_r$, based on $y_j$, the BG value at time $t_{j+h}$, for each $\vec{x}_j$:
\begin{gather*}
\CC_o = \left\{ \vec{x}_j \in \CC: 0 \leq y_j \leq 70 \right\} \ \textup{(hyp\textbf{o}glycemia)}; \\
\CC_e = \left\{ \vec{x}_j \in \CC: 70 <y_j \leq 180 \right\} \ \textup{(\textbf{e}uglycemia)}; \\
\CC_r = \left\{ \vec{x}_j \in \CC: 180 < y_j \leq 450 \right\} \ \textup{(hype\textbf{r}glycemia)},
\end{gather*}
with
$$ \CC_{\ell}^{\star} = \left\{ (\vec{x}_j, y_j) : \vec{x}_j \in \CC_{\ell} \right\}, \quad \ell \in \left\{ o,e,r\right\}. $$

In addition, we also classify each $\vec{x}_j \in \QQ$ as follows:
\begin{gather*}
\QQ_o = \left\{ \vec{x}_j \in \QQ: 0 \leq x_{j,d} \leq 70 \right\} \ \textup{(hyp\textbf{o}glycemia)}; \\
\QQ_e = \left\{ \vec{x}_j \in \QQ: 70 < x_{j,d} \leq 180 \right\} \ \textup{(\textbf{e}uglycemia)}; \\
\QQ_r = \left\{ \vec{x}_j \in \QQ: 180 <x_{j,d} \leq 450 \right\} \ \textup{(hype\textbf{r}glycemia)}.
\end{gather*}

Ideally, the classification of each $\vec{x}_j \in \QQ$ should also be done based on the value of $y_j$ (as was done for the classification of the training data $\CC$). However, since the values $y_j$ are only available for the training data and not for the test data, we use the next best available measurement for each $\vec{x}_j$, namely $x_{j,d}$, the BG value at time $t_j$.

\subsection{Prediction} 
\label{subsection-prediction}

Before making the BG predictions, it is necessary to scale the input readings so that the components of each $\vec{x}_j \in \PP$ are in $[-1/2, 1/2]$. This is done through the transformation
$$ \vec{x}_j \mapsto \frac{2\vec{x}_j - (M+m)}{2(M-m)}, $$
where $M := \max\{ x_{j,k}: \vec{x}_j \in \CC \}$ and $m := \min\{ x_{j,k}: \vec{x}_j \in \CC \}$.\\

With $\hat{F}_{n,\alpha}=\hat{F}_{n,\alpha}(Y,\vec{x}_j)$ defined as in \eqref{festimator} used with training data $Y$ and evaluated at a point $\vec{x}_j$, we are finally ready to compute the BG prediction for each $\vec{x}_j \in \QQ$ by
$$ f(\vec{x}_j) := \begin{cases}
\hat{F}_{n_o,\alpha_o}(\CC_o^{\star}, \vec{x}_j), & \vec{x}_j \in \QQ_o, \\
\hat{F}_{n_e,\alpha_e}(\CC_e^{\star}, \vec{x}_j), & \vec{x}_j \in \QQ_e, \\
\hat{F}_{n_r,\alpha_r}(\CC_r^{\star}, \vec{x}_j), & \vec{x}_j \in \QQ_r.
\end{cases} $$
The construction of the estimator $\hat{F}_{n,\alpha}$ involves several technical details regarding the classical Hermite polynomials. For ease of reading, the details of this construction are deferred to the appendix.

\subsection{Evaluation} 
\label{subsection-evaluation}

To evaluate the performance of the final output $f(\vec{x}_j),\ \vec{x}_j \in\QQ,$ we use the PRED-EGA mentioned in Section \ref{section-problem}. Specifically, a PRED-EGA grid is constructed by using comparisons of $f(\vec{x}_j)$ with the reference value $y_j$. This involves comparing
$$f(\vec{x}_j) \quad \textup{with} \quad y_j $$
as well as the rates of change
$$
\dfrac{f(\vec{x}_{j+1}) - f(\vec{x}_{j-1})}{2(t_{j+1} - t_{j-1})} \quad
\textup{with} \quad \dfrac{y_{j+1} -y_{j-1}}{2(t_{j+1} - t_{j-1})},
\label{eq-rateofchange}
$$
for all $\vec{x}_j \in \QQ$. Based on these comparisons, PRED-EGA classifies $f(\vec{x}_j)$ as being Accurate, Benign or Erroneous.\\

We repeat the entire process described in Subsections \ref{subsection-training} - \ref{subsection-evaluation} for a fixed number of trials, after which we report the average of the PRED-EGA grid placements, over all $\vec{x}_j \in \QQ$ and over all trials, as the final evaluation.\\

A summary of the method is given in Algorithm \ref{alg}.

\begin{algorithm}
	\SetKwInOut{Input}{input}
	\SetKwInOut{Output}{output}
	\Input{Time series $\left\{ {s}_p(t_j) \right\}$, $p \in P$, formatted as $\PP = \left\{ \vec{x}_j \right\}$ with \\$\vec{x}_j = \left( s_p(t_{j-d+1}), \cdots, s_p(t_j) \right)$ and $y_j = s_p(t_{j+h})$\\
		$d \in \N$ (specifies sampling horizon), $h \in \N$ (specifies prediction horizon)\\ $c \in (0,100)$ (percentage of data used for training)\\
		$n_o, n_e, n_r \in \N$, $\alpha_o, \alpha_e, \alpha_r \in (0,1]$ (parameters for function approximation) \\
		Let $C$ contain $c\%$ of patients from $P$ (drawn according to uniform prob. distr.) \; \\
		Set $\CC = \left\{ \vec{x}_j \right\}$ and $\CC^{\star} = \left\{ (\vec{x}_j, y_j) \right\}$ for all patients $p \in C$ \; \\
		Set $\QQ = \PP \setminus \CC $ \; }
	\Output{Prediction $f(\vec{x}_j)\approx s_p(t_{j+h})$ for $\vec{x}_j\in \mathcal{Q}$. }
	\BlankLine

	\SetKwProg{myproc}{}{}{}

		Set $\CC_o = \left\{ \vec{x}_j \in \CC: 0 \leq y_j \leq 70 \right\} $ \; \\
		Set $\CC_e = \left\{ \vec{x}_j \in \CC: 70 < y_j \leq 180 \right\} $ \; \\
		Set $\CC_r = \left\{ \vec{x}_j \in \CC: 180 < y_j \leq 450 \right\}$ \; \\
		Set $\CC_{\ell}^{\star} = \left\{ (\vec{x}_j, y_j) : \vec{x}_j \in \CC_{\ell} \right\}, \ \ell \in \left\{ o,e,r\right\}$ \; \\
		
		Set $\QQ_o = \left\{ \vec{x}_j \in \QQ: 0 \leq x_{j,d} \leq 70 \right\} $ \; \\
		Set $\QQ_e = \left\{ \vec{x}_j \in \QQ: 70 < x_{j,d} \leq 180 \right\} $ \; \\
		Set $\QQ_r = \left\{ \vec{x}_j \in \QQ: 180 < x_{j,d} \leq 450 \right\}$ \; \\
		
		Set $M = \max\{ x_{j,k}: \vec{x}_j \in \CC \}$ and $m = \min\{ x_{j,k}: \vec{x}_j \in \CC \}$ \; \\
		\For {$\vec{x}_j \in \PP$} {Rewrite $\vec{x}_j = \frac{2\vec{x}_j - (M+m)}{2(M-m)}$} \;

		\For {$\ell \in \left\{ o,e,r \right\}$} {\For {$\vec{x}_j \in \QQ_{\ell}$} {Compute $f(\vec{x}_j) = \hat{F}_{n_{\ell},\alpha_{\ell}}(\CC_{\ell}^{\star},\vec{x}_j)$} \;
		}  \;

	\caption{Deep Network for BG prediction}
	\label{alg}
\end{algorithm}

\section{Results and discussion}
\label{section-results}

As mentioned in Section \ref{section-problem}, we apply our method to data provided by the DirecNet Central Laboratory. Time series for the 25 patients in data set D and the 25 patients in data set J that contain the largest number of BG measurements are considered. These specific data sets were obtained to study the performance of CGM devices in children with Type I diabetes; as such, all of the patients are less than 18 years old. Summary statistics of the two data sets are provided in Table \ref{table-data}. Our method is a general purpose algorithm, where these details do not play any significant role, except in affecting the outcome of the experiments.\\

We provide results obtained by implementing our method in MATLAB, as described in Algorithm \ref{alg}. For our implementation, we employ a sampling horizon $t_{j}-t_{j-d+1}$ of 30 minutes ($d=7$), 50\% training data ($c=50$) (which is comparable to approaches followed in for example \cite{reifman2007predictive}) and a total of 100 trials ($T=100$). For all the experiments, the function approximation parameters $n_o, n_e$ and $n_r$ referenced in Algorithm \ref{alg} were chosen in the range $\{3,\ldots,7\}$ with $\alpha_o=\alpha_e=\alpha_r = 1$. We provide results for prediction horizons $t_{j+m} - t_j$ of 30 minutes ($h=6$), 60 minutes ($h=12$) and 90 minutes ($h=18$) (again, comparable to prediction horizons in for example \cite{reifman2007predictive,seo2019machine,zhu2020dilated}). After testing our method on all 25 patients in each data set separately, the average PRED-EGA scores (in percent) are displayed in Tables \ref{table-A} and \ref{table-C}. \\

For comparison, Tables \ref{table-A} and \ref{table-C} also display predictions on the same data sets using four other methods:
\begin{enumerate}[(i)]
	\item MATLAB's built-in Deep Learning Toolbox. 
	We used a network architecture consisting of three fully connected convolutional and output layers and appropriate input, normalization, activation, and averaging  layers. As before, we used 50\% training data and averaged the experiment over 100 trials, for prediction windows of 30 minutes, 60 minutes and 90 minutes and data sets D and J.
	\item The diffusion geometry approach, based on the eigen-decomposition of the Laplace-Beltrami operator, we followed in our 2017 paper \cite{mhas_sergei_maryke_diabetes2017}. As mentioned in Section \ref{section-intro}, this approach can be classified as semi-supervised learning, where we need to have all the data points $\vec{x}_j$ available in advance. Because of this requirement, this approach cannot be used  for prediction based on data that was not included in the original data set we worked with. Indeed, this is one of the main motivations for our current method, which is not dependent on this requirement. Nevertheless, we include a comparison of the diffusion geometry approach with our current method for greater justification of the accuracy of our current method. Again, we used 50\% training data and averaged the experiment over 100 trials, for prediction windows of 30 minutes, 60 minutes and 90 minutes and data sets D and J.
	\item The Legendre polynomial prediction method followed in our 2013 paper \cite{mhaskar2013filtered}. In this context, the main mathematical problem can be summarized as that of estimating the derivative of a function at the end point of an interval, based on measurements of the function in the past. 
	An important difference between our current method and the Legendre polynoimal approach is that with the latter, learning is based on the data for each patient separately (each patient forms a data set in itself, as explained in Section \ref{section-intro}). Therefore, the method is philosophically different from ours. We nevertheless include a comparison for prediction windows of 30 minutes, 60 minutes and 90 minutes and data sets D and J.
	\item The Fully Adaptive Regularized Learning (FARL) approach in the 2012 paper \cite{naumova2012meta}, which uses meta-learning to choose the kernel and regularization parameters adaptively (briefly described in Section \ref{section-priorwork}). As explained in \cite{naumova2012meta}, a strong suit of FARL is that it is expected to be portable from patient to patient without any readjustment. Therefore, the results reported below were obtained by implementing the code and parameters used in \cite{naumova2012meta} directly on data sets D and J for prediction windows of 30 minutes, 60 minutes and 90 minutes, with no adjustments for the kernel and regularization parameters.
	As explained earlier, this method is also not directly compared to our theoretically well founded method. 
\end{enumerate}

\begin{figure}
	\centering
	\includegraphics[width=0.75\linewidth]{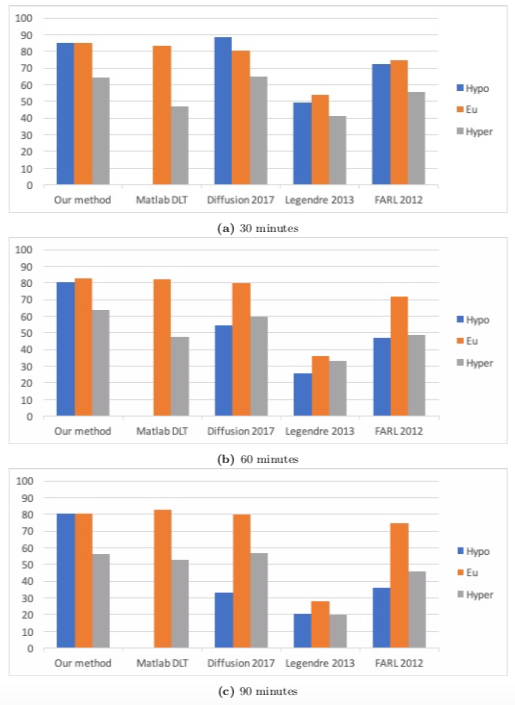}
	\caption{Percentage accurate predictions and predictions with benign consequences in all three BG ranges (blue: hypoglycemia; orange: euglycemia; grey: hyperglycemia) for \textbf{dataset D}, for all five methods for different prediction windows (30 minutes, 60 minutes and 90 minutes).}
	\label{fig-c}
\end{figure}

\begin{figure}
	\centering
	\includegraphics[width=0.75\linewidth]{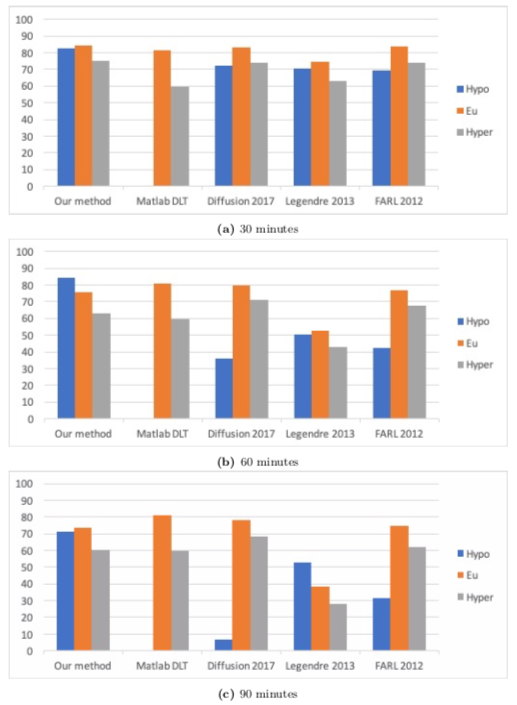}
	\caption{Percentage accurate predictions and predictions with benign consequences in all three BG ranges (blue: hypoglycemia; orange: euglycemia; grey: hyperglycemia) for \textbf{dataset J}, for all five methods for different prediction windows (30 minutes, 60 minutes and 90 minutes).}
	\label{fig-d}
\end{figure}

The percentage accurate predictions and predictions with benign consequences in all three BG ranges, as obtained using all five methods, for a 30 minute, 60 minute and 90 minute prediction window, are displayed visually in Figures \ref{fig-c}-\ref{fig-d}.\\

By comparing the percentages in Tables \ref{table-A}-\ref{table-C} and the bar heights in Figures \ref{fig-c}-\ref{fig-d}, it is clear that our method far outperforms the competitors in the hypoglycemic range, except for the 30 minute prediction on data set D, where the diffusion geometry approach is slightly more accurate -- but it is important to remember that the diffusion geometry approach has all the data points available from the outset. The starkest difference is between our method and Matlab's Deep Learning Toolbox -- the latter achieves less than 1.5\% accurate and benign consequence predictions regardless of the size of the prediction window or data set. There is also a marked difference between our method and the Legendre polynomial approach, especially for longer prediction horizons. Our method achieves comparable or higher accuracy in the hyperglycemic range, except for the 60 and 90 minute predictions on data set J, where the diffusion geometry and FARL approaches achieve more accurate predictions. The methods display comparable accuracy in the euglycemic range, except for the 60 and 90 minute predictions on dataset J, where Matlab's toolbox, the diffusion geometry approach and the FARL method achieve higher accuracy.\\

\begin{figure}
	\centering
	\includegraphics[width=\textwidth]{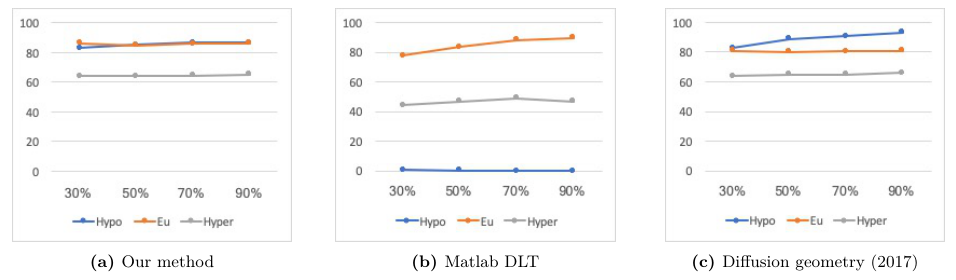}
	\caption{Percentage accurate predictions and predictions with benign consequences in all three BG ranges (blue: hypoglycemia; orange: euglycemia; grey: hyperglycemia) with a 30 minute prediction window on \textbf{dataset D}, for different training set sizes (30\%, 50\%, 70\% and 90\%). }
		\label{fig-f}
\end{figure}

\begin{figure}
	\centering
	\includegraphics[width=\textwidth]{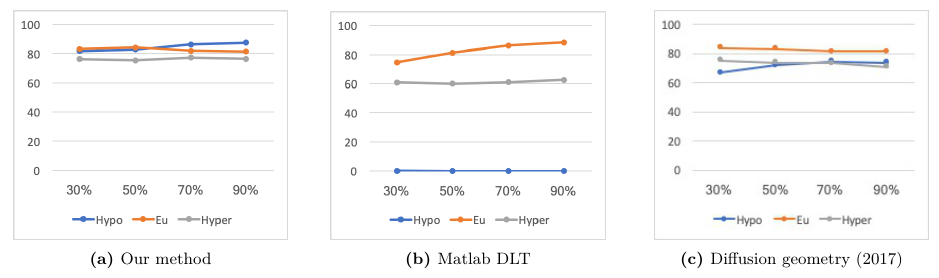}
	\caption{Percentage accurate predictions and predictions with benign consequences in all three BG ranges (blue: hypoglycemia; orange: euglycemia; grey: hyperglycemia) with a 30 minute prediction window on \textbf{dataset J}, for different training set sizes (30\%, 50\%, 70\% and 90\%). }
		\label{fig-g}
\end{figure}

Tables \ref{table-E}-\ref{table-F} display the results when testing those methods that are dependent on training data selection (that is, our current method, the Deep Learning Toolbox and the diffusion geometry approach in our 2017 paper) on different training set sizes (namely, 30\%, 50\%, 70\% and 90\%). For these experiments, we used a fixed prediction window of 30 minutes, and training data was drawn randomly according to a uniform probability distribution in each case. While the Deep Learning Toolbox and diffusion geometry approach sometimes perform slightly better than our method in the euglycemic and hypoglycemic ranges, respectively, our method consistently outperforms both of these in the other two blood glucose ranges. The results are displayed visually as well in Figures \ref{fig-f}-\ref{fig-g}.\\

As mentioned in Section \ref{section-intro}, a feature of our method is that it does not require any information about the data manifold other than its dimension. So, in principle, one could ``train'' on one data set and apply the resulting model to another data set taken from a different manifold. We illustrate this by constructing our model based on one of the data sets D or J and testing it on the other data set. Building on this idea, we also demonstrate the accuracy of our method as compared to MATLAB's Deep Learning Toolbox when trained on data set D and applied to perform prediction on data set J, and vice versa, which is the only other method that is directly comparable with the philosophy behind our method. These results are reported in Tables \ref{table-B} and \ref{table-D}. It is clear that we are successful in our goal of training on one data set and predicting on a different data set. The percentage accurate predictions and predictions with benign consequences when training and testing on different datasets are comparable to the cases when we are training and testing on the same dataset; in fact, we obtain even better results in the hypoglycemic BG range when training and testing across different datasets.\\

\begin{figure}
	
	\centering
		\includegraphics[width=\textwidth]{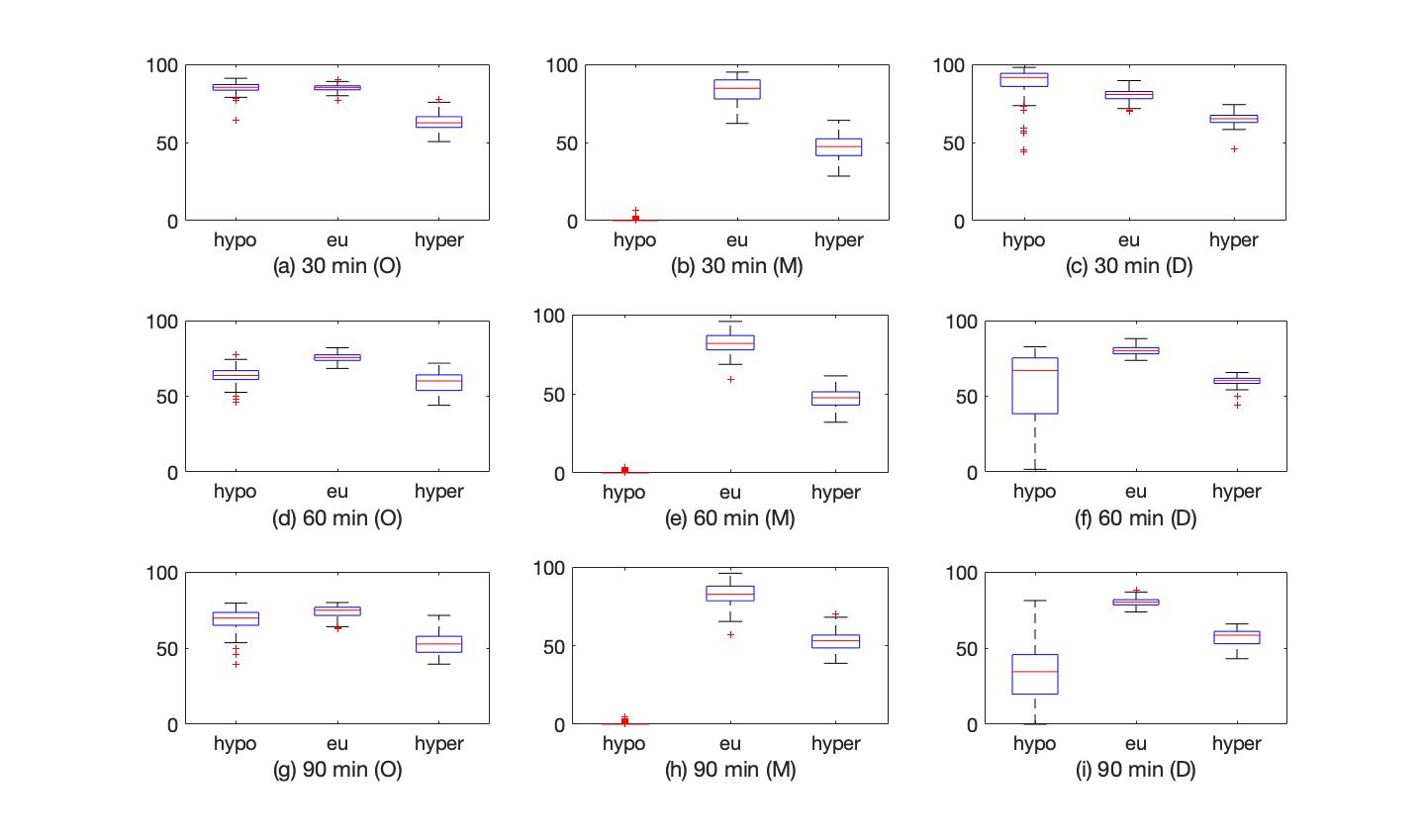}
	\caption{Boxplot for the 100 experiments conducted with 50\% training data for each prediction method (O = our method, M = MATLAB Deep Learning Toolbox, D = diffusion geometry approach) with a 30 minute (top), 60 minute (middle) and 90 minute (bottom) prediction horizon and \textbf{dataset D}. Each of the graphs show the percentage accurate predictions in the hypoglycemic range (left), euglycemic range (middle) and hyperglycemic range (right).}
	\label{fig-a}
\end{figure}

\begin{figure}
	
	\centering
	\includegraphics[width=\textwidth]{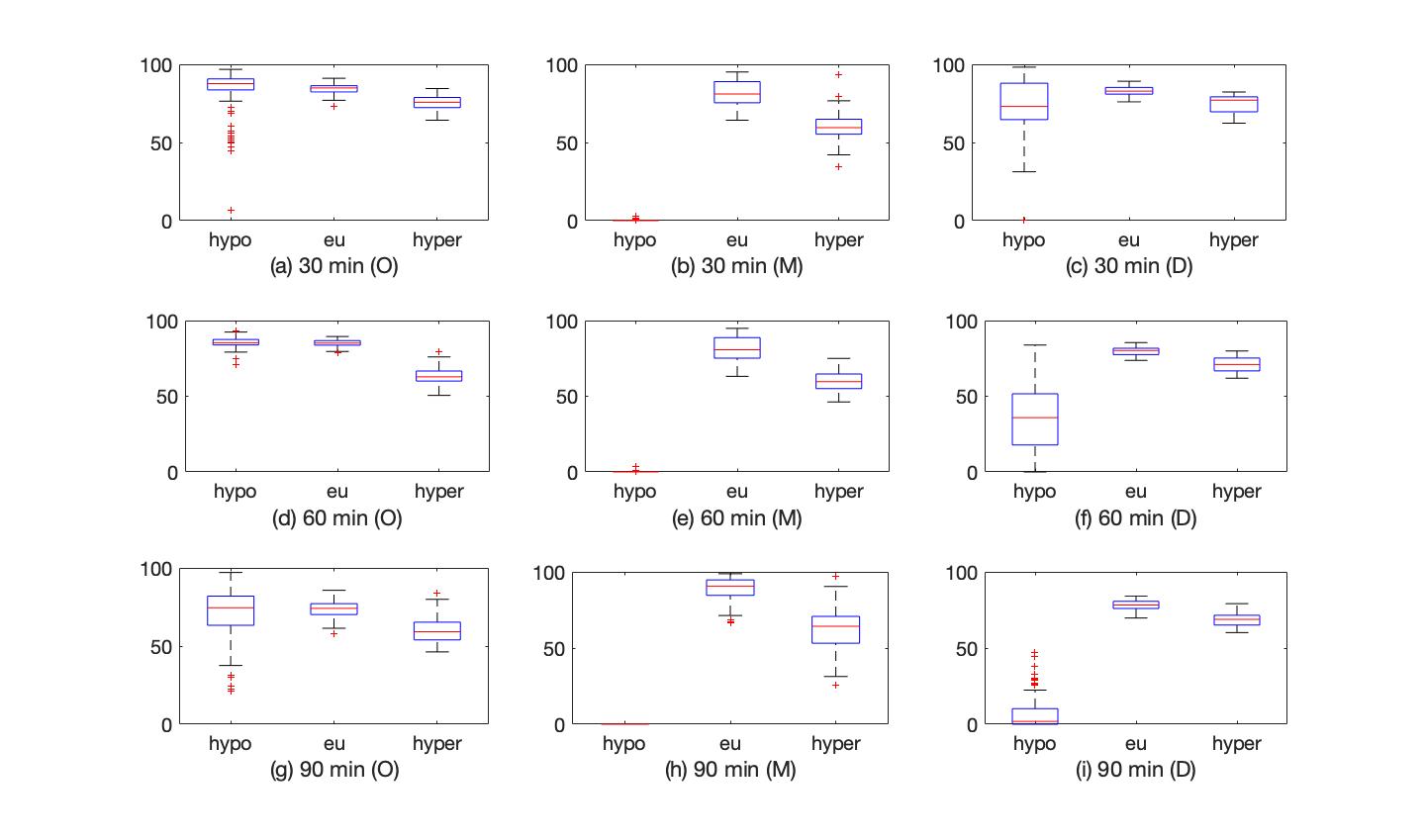}
	
	\caption{Boxplot for the 100 experiments conducted with 50\% training data for each prediction method (O = our method, M = MATLAB Deep Learning Toolbox, D = diffusion geometry approach) with a 30 minute (top), 60 minute (middle) and 90 minute (bottom) prediction horizon and \textbf{dataset J}. Each of the graphs show the percentage accurate predictions in the hypoglycemic range (left), euglycemic range (middle) and hyperglycemic range (right).}
	\label{fig-b}
\end{figure}

Figures \ref{fig-a} and \ref{fig-b} display box plots for the 100 trials for the methods dependent on training data selection (that is, our current method, the Deep Learning Toolbox and the 2017 diffusion geometry approach) and prediction windows using data sets D and J, respectively. Our method performs fairly consistently over different trials.\\

\begin{table}
	\begin{center}
		\begin{tabular}{c| c c c | c c c | ccc}
			\hline
			& \multicolumn{3}{|c|}{Hypoglycemia:} & \multicolumn{3}{|c|}{Euglycemia:} & \multicolumn{3}{|c}{Hyperglycemia:} \\
			& \multicolumn{3}{|c|}{BG $\leq 70$ (mg/dL)} & \multicolumn{3}{|c|}{BG $70 - 180$ (mg/dL)} & \multicolumn{3}{|c}{BG $> 180$ (mg/dL)} \\
			& Acc. & Benign & Error & Acc. & Benign & Error & Acc. & Benign & Error\\
			\hline
			\multicolumn{10}{l}{\textbf{30 min prediction horizon:}} \\ \hline
			Our method & 85.05 &	13.34 &	1.61 &	84.96 &	11.93 &	3.11 &	64.18 &	22.89 &	12.93 \\ \hline
			Matlab DL Toolbox & 0.24 &	0.26 &	99.50 &	83.42 &	13.94 &	2.65 &	47.00 &	16.73 &	36.27 \\ \hline
			Diffusion geometry (2017) & 88.72 &	4.49 &	6.79 &	80.32 &	17.36 &	2.32 &	64.88 &	21.90 &	13.22 \\ \hline
			Legendre polynomials (2013) & 49.49 &	25.85 &	24.66 &	53.88 &	37.28 &	8.84 &	41.40 &	34.90 &	23.70 \\ \hline
			FARL (2012)& 72.42 &	8.80 &	18.78 &	74.54 &	18.79 &	6.67 &	55.45 &	26.42 &	18.13 \\ \hline
			\multicolumn{10}{l}{\textbf{60 min prediction horizon:}} \\ \hline
			Our method & 80.44 &	12.33 &	7.23 &	82.78 &	11.59 &	5.63 &	63.83 &	20.92 &	15.25 \\ \hline
			Matlab DL Toolbox & 0.39 &	0.62 &	98.99 &	82.30 &	14.70 &	3.00 &	47.48 &	18.70 &	33.82 \\ \hline
			Diffusion geometry (2017) & 54.30 &	4.72 &	40.98 &	79.99 &	17.29 &	2.72 &	59.78 &	21.13 &	19.09 \\ \hline
			Legendre polynomials (2013) & 25.66 &	34.54 &	39.80 &	36.20 &	49.44 &	14.36 &	33.27 &	32.21 &	34.52 \\ \hline
			FARL (2012)& 47.11 &	6.80 &	46.10 &	71.62 &	21.29 &	7.09 &	48.57 &	23.91 &	27.51 \\ \hline
			\multicolumn{10}{l}{\textbf{90 min prediction horizon:}} \\ \hline
			Our method & 80.60 &	10.05 &	9.35 &	80.19 &	9.96 &	9.85 &	56.07 &	15.44 &	28.49 \\ \hline
			Matlab DL Toolbox & 0.32 &	0.38 &	99.30 &	83.01 &	14.26 &	2.73 &	53.05 &	13.77 &	33.18 \\ \hline
			Diffusion geometry (2017) & 33.07 &	4.42 &	62.51 &	80.12 &	15.87 &	4.01 &	56.73 &	20.84 &	22.43 \\ \hline
			Legendre polynomials (2013) & 20.67 &	39.28 &	40.05 &	27.98 &	51.64 &	20.38 &	20.00 &	30.94 &	49.06 \\ \hline
			FARL (2012)& 35.98 &	2.53 &	61.49 &	74.41 &	16.97 &	8.62 &	46.17 &	26.69 &	27.14 \\ \hline
		\end{tabular}
	\end{center}
	\caption{Average PRED-EGA scores (in percent) for different prediction horizons on \textbf{dataset D}.}
	\label{table-A}
\end{table}

\begin{table}
		\begin{center}
			\begin{tabular}{c| c c c | c c c | ccc}
				\hline
				& \multicolumn{3}{|c|}{Hypoglycemia:} & \multicolumn{3}{|c|}{Euglycemia:} & \multicolumn{3}{|c}{Hyperglycemia:} \\
				& \multicolumn{3}{|c|}{BG $\leq 70$ (mg/dL)} & \multicolumn{3}{|c|}{BG $70 - 180$ (mg/dL)} & \multicolumn{3}{|c}{BG $> 180$ (mg/dL)} \\
				& Acc. & Benign & Error & Acc. & Benign & Error & Acc. & Benign & Error\\
				\hline
				\multicolumn{10}{l}{\textbf{30 min prediction horizon:}} \\ \hline
				Our method & 82.80 & 6.60 &	10.60 &	84.30 &	10.90 &	4.80 &	75.33 &	11.48 &	13.19 \\ \hline
				Matlab DL Toolbox & 0.06 &	0.06 &	99.88 &	81.40 &	14.50 &	4.10 &	59.87 &	8.85 &	31.28 \\ \hline
				Diffusion geometry (2017) & 72.09 &	1.52 &	26.39 &	83.12 &	14.74 &	2.14 &	73.91 &	12.03 &	14.06 \\ \hline
				Legendre polynomials (2013) & 70.68 &	11.75 &	17.57 &	74.48 &	19.67 &	5.85 &	62.82 &	26.08 &	11.10 \\ \hline
				FARL (2012)& 69.61 &	7.98 &	22.41 &	83.87 &	12.31 &	3.82 &	73.95 &	16.43 &	9.62 \\ \hline
				\multicolumn{10}{l}{\textbf{60 min prediction horizon:}} \\ \hline
				Our method & 84.27 &	5.66 &	10.07 &	75.43 &	11.11 &	13.46 &	63.13 &	8.49 &	28.38 \\ \hline
				Matlab DL Toolbox & 0.06 &	0.13 &	99.81 &	81.12 &	14.64 &	4.24 &	59.83 &	8.21 &	31.96 \\ \hline
				Diffusion geometry (2017) & 35.89 &	0.87 &	63.24 &	79.50 &	14.96 &	5.54 &	71.37 &	12.53 &	16.10 \\ \hline
				Legendre polynomials (2013) & 50.16 &	22.65 &	27.19 &	52.44 &	34.36 &	13.20 &	42.89 &	36.08 &	21.03 \\ \hline
				FARL (2012)& 42.28 &	1.61 &	56.11 &	76.74 &	14.48 &	8.78 &	67.56 &	16.75 &	15.69 \\ \hline
				\multicolumn{10}{l}{\textbf{90 min prediction horizon:}} \\ \hline
				Our method & 71.07 &	5.94 &	22.99 &	73.64 &	9.00 &	17.36 &	60.28 &	6.30 &	33.42 \\ \hline
				Matlab DL Toolbox & 0.10 &	0.50 &	99.85 &	81.09 &	14.75 &	4.16 &	59.47 &	9.49 &	31.04 \\ \hline
				Diffusion geometry (2017) & 6.84 &	1.03 &	92.13 &	77.85 &	13.93 &	8.22 &	68.29 &	13.04 &	18.67 \\ \hline
				Legendre polynomials (2013) & 52.58 &	15.71 &	31.71 &	38.46 &	40.44 &	21.10 &	27.92 &	37.13 &	34.95 \\ \hline
				FARL (2012)& 31.58 &	1.23 &	67.18 &	74.96 &	13.10 &	11.95 &	61.96 &	14.06 &	23.98 \\ \hline
			\end{tabular}
		\end{center}
	\caption{Average PRED-EGA scores (in percent) for different prediction horizons on \textbf{dataset J}.}
	\label{table-C}
\end{table}

\begin{table}
	\begin{center}
		\begin{tabular}{c| c c c }
			\hline
			& Hypoglycemia: & Euglycemia: & Hyperglycemia: \\
			& BG $\leq 70$ (mg/dL) & BG $70 - 180$ (mg/dL) & BG $> 180$ (mg/dL) \\
			\hline
			\multicolumn{4}{l}{\textbf{30\% training data:}} \\ \hline
			Our method & 83.02 &	86.15 &	64.14  \\ \hline
			Matlab DL Toolbox & 0.88 &	77.89 &	44.50  \\ \hline
			Diffusion geometry (2017) & 82.56 &	80.83 &	63.89  \\ \hline
			\multicolumn{4}{l}{\textbf{50\% training data:}} \\ \hline
			Our method & 85.05 &	84.96 &	64.18  \\ \hline
			Matlab DL Toolbox & 0.24 &	83.42 &	47.00  \\ \hline
			Diffusion geometry (2017) & 88.72 &	80.32 &	64.88  \\ \hline
			\multicolumn{4}{l}{\textbf{70\% training data:}} \\ \hline
			Our method & 86.62 &85.79 &	64.47  \\ \hline
			Matlab DL Toolbox & 0.06 &	88.14 &	48.94  \\ \hline
			Diffusion geometry (2017) & 90.72 &	80.34 &	64.99  \\ \hline
			\multicolumn{4}{l}{\textbf{90\% training data:}} \\ \hline
			Our method & 86.77 &	86.34 &	65.20  \\ \hline
			Matlab DL Toolbox & 0.09 &	89.71 &	46.97  \\ \hline
			Diffusion geometry (2017) & 93.22 &	80.87 &	65.84  \\ \hline
		\end{tabular}
	\end{center}
	\caption{Percentage accurate predictions for different training set sizes and a 30 min prediction horizon on \textbf{dataset D}.}
	\label{table-E}
\end{table}

\begin{table}
	\begin{center}
		\begin{tabular}{c| c c c }
			\hline
			& Hypoglycemia: & Euglycemia: & Hyperglycemia: \\
			& BG $\leq 70$ (mg/dL) & BG $70 - 180$ (mg/dL) & BG $> 180$ (mg/dL) \\
			\hline
			\multicolumn{4}{l}{\textbf{30\% training data:}} \\ \hline
			Our method & 81.79 &	83.34 &	76.12  \\ \hline
			Matlab DL Toolbox & 0.07 &	74.75 &	60.81  \\ \hline
			Diffusion geometry (2017) & 67.02 &	83.99 &	75.33  \\ \hline
			\multicolumn{4}{l}{\textbf{50\% training data:}} \\ \hline
			Our method & 82.80 &	84.30 &	75.33  \\ \hline
			Matlab DL Toolbox & 0.06 &	81.40 &	59.87  \\ \hline
			Diffusion geometry (2017) & 72.09 &	83.12 &	73.91  \\ \hline
			\multicolumn{4}{l}{\textbf{70\% training data:}} \\ \hline
			Our method & 86.40 &	81.89 &	77.21  \\ \hline
			Matlab DL Toolbox & 0.00 &	86.26 &	60.93  \\ \hline
			Diffusion geometry (2017) & 74.45 &	81.28 &	73.47  \\ \hline
			\multicolumn{4}{l}{\textbf{90\% training data:}} \\ \hline
			Our method & 87.67 &	81.43 &	76.40  \\ \hline
			Matlab DL Toolbox & 0 &	88.30 &	62.45  \\ \hline
			Diffusion geometry (2017) & 73.90 &	81.33 &	70.98  \\ \hline
		\end{tabular}
	\end{center}
	\caption{Percentage accurate predictions for different training set sizes and a 30 min prediction horizon on \textbf{dataset J}.}
		\label{table-F}
\end{table}

\begin{table}
	\begin{center}
		\begin{tabular}{c| c c c | c c c | ccc}
			\hline
			& \multicolumn{3}{|c|}{Hypoglycemia:} & \multicolumn{3}{|c|}{Euglycemia:} & \multicolumn{3}{|c}{Hyperglycemia:} \\
			& \multicolumn{3}{|c|}{BG $\leq 70$ (mg/dL)} & \multicolumn{3}{|c|}{BG $70 - 180$ (mg/dL)} & \multicolumn{3}{|c}{BG $> 180$ (mg/dL)} \\
			& Acc. & Benign & Error & Acc. & Benign & Error & Acc. & Benign & Error\\
			\hline
			\multicolumn{10}{l}{\textbf{30 min prediction horizon:}} \\ \hline
			Our method & 85.85 &	12.23 &	1.92 &	85.51 &	11.48 &	3.01 &	54.95 &	24.21 &	20.84 \\ \hline
			Matlab DL Toolbox & 0.01 &	0.00 &	99.99 &	90.91 &	8.09 &	1.01 &	50.43 &	14.99 &	34.58 \\ \hline
			\multicolumn{10}{l}{\textbf{60 min prediction horizon:}} \\ \hline
			Our method & 85.05 &	12.30 &	2.65 &	79.86 &	10.59 &	9.55 &	45.73 &	18.52 &	35.75 \\ \hline
			Matlab DL Toolbox & 0.00 &	0.01 &	99.99 &	90.57 &	8.39 &	1.03 &	52.37 &	16.08 &	31.55 \\ \hline
			\multicolumn{10}{l}{\textbf{90 min prediction horizon:}} \\ \hline
			Our method & 78.01 &	9.41 &	12.58 &	79.65 &	10.16 &	10.19 &	43.18 &	12.36 &	44.46 \\ \hline
			Matlab DL Toolbox & 0.00 &	0.01 &	99.99 &	89.83 &	8.74 &	1.43 &	56.35 &	12.58 &	31.07 \\ \hline
		\end{tabular}
	\end{center}
	\caption{Average PRED-EGA scores (in percent) for different prediction horizons on \textbf{dataset D}, with \textbf{training data selected from dataset J}.}
	\label{table-B}
\end{table}

\begin{table}
	\begin{center}
		\begin{tabular}{c| c c c | c c c | ccc}
			\hline
			& \multicolumn{3}{|c|}{Hypoglycemia:} & \multicolumn{3}{|c|}{Euglycemia:} & \multicolumn{3}{|c}{Hyperglycemia:} \\
			& \multicolumn{3}{|c|}{BG $\leq 70$ (mg/dL)} & \multicolumn{3}{|c|}{BG $70 - 180$ (mg/dL)} & \multicolumn{3}{|c}{BG $> 180$ (mg/dL)} \\
			& Acc. & Benign & Error & Acc. & Benign & Error & Acc. & Benign & Error\\
			\hline
			\multicolumn{10}{l}{\textbf{30 min prediction horizon:}} \\ \hline
			Our method & 82.18 &	17.59 &	0.23 &	79.72 &	14.04 &	6.24 &	74.19 &	16.30 &	9.51 \\ \hline
			Matlab DL Toolbox & 0.01 &	0.00 &	99.99 &	89.16 &	9.44 &	1.40 &	60.47 &	4.62 &	34.91 \\ \hline
			\multicolumn{10}{l}{\textbf{60 min prediction horizon:}} \\ \hline
			Our method & 88.08 &	10.07 &	1.85 &	73.50 &	14.41 &	12.09 &	63.90 &	11.64 &	24.46 \\ \hline
			Matlab DL Toolbox & 0.05 &	0.01 &	94.94 &	89.79 &	9.05 &	1.16 &	60.94 &	4.20 &	34.86 \\ \hline
			\multicolumn{10}{l}{\textbf{90 min prediction horizon:}} \\ \hline
			Our method & 74.79 &	10.36 &14.85 &	67.83 &	13.20 &	18.97 &	57.70 &	10.55 &	31.75 \\ \hline
			Matlab DL Toolbox & 0.09 &	0.14 &	99.77 &	89.38 &	9.38 &	1.24 &	58.85 &	4.98 &	36.17 \\ \hline
		\end{tabular}
	\end{center}
	\caption{Average PRED-EGA scores (in percent) for different prediction horizons on \textbf{dataset J}, with \textbf{training data selected from dataset D}.}
	\label{table-D}
\end{table}

\section{Conclusions}
\label{section-conclusions}

In this paper, we demonstrate a direct method to approximate functions on unknown manifolds \cite{mhaskar2019deep} in the context of BG prediction for diabetes management. 
The results are evaluated using the state-of-the-art PRED-EGA grid \cite{sivananthan2011assessment}. 
Unlike classical manifold learning, our method yields a model for the BG prediction on data not seen during training, even for test data which is drawn from a different distribution from the training data.
Our method outperforms other methods in the literature in 30-minute, 60-minute and 90-minute predictions in the hypoglycemic and hyperglycemic BG ranges, and is especially useful for BG prediction for patients that are not included in the training set.

\section*{Acknowledgments}
The research of HNM is  supported in part by NSF grant DMS 2012355 and ARO grant W911NF2110218.
The research by SVP has been partly supported by BMK, BMDW, and the Province of Upper Austria in the frame of the COMET Programme managed by FFG in the COMET Module S3AI.

\bibliographystyle{plain}
\bibliography{bib_bg}


\appendix
\renewcommand{\theequation}{\Alph{section}.\hindu{equation}}
\section{Theoretical aspects}
\label{hermitesect}

In this section, we describe the theoretical background behind the estimator $\widehat{F}_{n,\alpha}(\Gamma,\circ)$ used in this paper to predict the BG level.
We focus only on the essential definitions, referring to \cite{mhaskar2019deep} for details.

Let $d\ge q\ge 1$ be integers, $\XX$ be a $q$ dimensional, compact, connected,  sub-manifold of $\RR^d$ (without boundary), with geodesic distance $\rho$ and volume measure $\mu^*$, normalized so that $\mu^*(\XX)=1$.
The operator $\widehat{F}_{n,\alpha}$ does not require any knowledge of the manifold other than $q$. 
Our construction is based on the classical Hermite polynomials which are best introduced by the recurrence relations \eqref{eq-recurrence}.
The orthonormalized Hermite polynomial $h_k$ of degree $k$ is defined recursively by
\begin{align}
h_k(x)&:=\sqrt{\frac{2}{k}}xh_{k-1}(x)-\sqrt{\frac{k-1}{k}}h_{k-2}(x), \qquad k=2,3,\cdots,
\\
h_0(x)&:=\pi^{-1/4},\ h_1(x):=\sqrt{2}\pi^{-1/4}x.
\label{eq-recurrence}
\end{align}
We write 
\be\label{uni_psi_def}
\psi_k(x):=h_k(x)\exp(-x^2/2), \qquad x\in\RR,\ k\in\ZZ_+.
\ee 

The functions $\{\psi_k\}_{k=0}^\infty$ are an orthonormal set with respect to the Lebesgue measure:
$$
\int_\RR \psi_k(x)\psi_j(x)dx=\begin{cases}
1, & \mbox{ if $k=j$,}\\
0, &\mbox{ otherwise}.
\end{cases}
$$
In the sequel, we fix an infinitely differentiable function $H :[0,\infty)\to [0,1]$, such that $H(t)=1$ if $0\le t\le 1/2$, and $H(t)=0$ if $t\ge 1$.
We define for $x\in\RR$, $m\in\ZZ_+$:
\be\label{fastproj}
\mathcal{P}_{m,q}(x):=\begin{cases}
\disp\pi^{-1/4} (-1)^m\frac{\sqrt{(2m)!}}{2^m m!}\psi_{2m}(x), &\mbox{ if $q=1$,}\\[2ex]
\disp \frac{1}{\pi^{(2q-1)/4}\Gamma((q-1)/2)}\sum_{\ell=0}^m (-1)^\ell\frac{\Gamma((q-1)/2+m-\ell)}{(m-\ell)!}  \frac{\sqrt{(2\ell)!}}{2^\ell \ell!}\psi_{2\ell}(x), &\mbox{ if $q\ge 2$,}
\end{cases}
\ee
and the kernel $\widetilde{\Phi}_{n,q}$ for $x\in\RR$, $n\in\ZZ_+$ by
\be\label{fastkerndef}
\widetilde{\Phi}_{n,q}(x):=\sum_{m=0}^{\lfloor n^2/2\rfloor}H\left(\frac{\sqrt{2m}}{n}\right)\mathcal{P}_{m,q}(x).
\ee

We assume here a noisy data of the form $(\y,\epsilon)$, with a joint probability distribution $\tau$ and assume further that  the marginal distribution of $\y$ with respect to $\tau$ has the form $d\nu^*=f_0d\mu^*$ for some $f_0\in C(\XX)$. 
In place of $f(\y)$, we consider a noisy variant $\mathcal{F}(\y,\epsilon)$, and denote
\be\label{Fdef} 
f(\y):=\mathbb{E}_\tau(\mathcal{F}(\y,\epsilon)|\y).
\ee

\begin{rem}\label{rem:offmanifold}
{\rm
In practice, the data may not lie on a manifold, but it is reasonable to assume that it lies on a tubular neighborhood of the manifold. 
Our notation accommodates this - if $\z$ is a point in a neighborhood of $\XX$, we may view it as a perturbation  of a point $\y\in\XX$, so that the noisy value of the target function is $\mathcal{F}(\y,\epsilon)$, where $\epsilon$ encapsulate the noise in both the $\y$ variable and the value of the target function.
\qed}
\end{rem}
Our approximation process is simple: given by
\be\label{festimator}
\widehat{F}_{n,\alpha}(Y;\x):=\frac{n^{q(1-\alpha)}}{M}\sum_{j=1}^M \mathcal{F}(\y_j,\epsilon_j) \tilde\Phi_{n,q}(n^{1-\alpha}|\x-\y_j|_{2,d}), \qquad \x\in\RR^d,
\ee
where $0<\alpha\le 1$.

The theoretical underpinning of our method is described by Theorem~\ref{theo:manifoldprob}, and in particular, by Corollary~\ref{cor:manifold_approx_prob}, describing the convergence properties of the estimator.
In order to state these results,
we need a smoothness class $W_\gamma(\XX)$, representing the assumptions necessary to guarantee the rate of convergence of our estimator to the target function. It is beyond the scope of this paper to describe the details of this smoothness class; an interested reader will find them in \cite{mhaskar2019deep}. 
We note that the definition of the estimator in \eqref{festimator} is \textit{universal}, and does not require any assumptions. 
The assumptions are needed only to gurantee the right rates of convergence.

\begin{theorem}\label{theo:manifoldprob}
Let $\gamma>0$, 
$\tau$ be a probability distribution on $\XX\times \Omega$ for some sample space $\Omega$ such  the marginal distribution of $\tau$ restricted to $\XX$ is absolutely continuous with respect to $\mu^*$ with density $f_0\in W_\gamma(\XX)$.
We assume that
\be\label{ballmeasure}
\sup_{\x\in\XX, r>0}\frac{\mu^*(\BB(\x,r))}{r^q} \le c.
\ee
Let $\mathcal{F} : \XX\times \Omega\to \RR$ be a bounded function,  $f$  defined by \eqref{Fdef} be in $W_\gamma(\XX)$, the probability density $f_0\in W_\gamma(\XX)$. 
Let $M\ge 1$, $Y=\{(\y_1,\epsilon_1),\cdots,(y_M,\epsilon_M)\}$ be a set of random samples chosen i.i.d. from $\tau$.
If
\be\label{alphacond}
0<\alpha<\frac{4}{2+\gamma}, \qquad \alpha\le 1,
\ee
then  for every $n\ge 1$, $0<\delta<1$ and $M\ge n^{q(2-\alpha)+2\alpha\gamma}\sqrt{\log (n/\delta)}$, we have with $\tau$-probability $\ge 1-\delta$:
\be\label{locprobest}
\left\|\widehat{F}_{n,\alpha}(Y;\circ)-ff_0\right\|_\XX\le c_1\frac{\sqrt{\|f_0\|_\XX}\|\mathcal{F}\|_{\XX\times \Omega}+\|ff_0\|_{W_\gamma(\XX)}}{n^{\alpha\gamma}}.
\ee
\end{theorem}

\begin{cor}\label{cor:manifold_approx_prob}
With the set-up as in Theorem~\ref{theo:manifoldprob}, let
 $f_0\equiv 1$ (i.e., the marginal distribution of $\y$ with respect to $\tau$ is $\mu^*$). Then we have with $\tau$-probability $\ge 1-\delta$:
\be\label{locprobest_approx}
\left\|\widehat{F}_{n,\alpha}(Y;\circ)-f\right\|_\XX\le c_1\frac{\|\mathcal{F}\|_{\XX\times \Omega}+\|f\|_{W_\gamma(\XX)}}{n^{\alpha\gamma}}.
\ee
\end{cor}


\end{document}